\documentclass[10pt,twocolumn,letterpaper]{article}

\usepackage[pagenumbers]{cvpr} %

\usepackage{graphicx}
\usepackage{amsmath}
\usepackage{amssymb}
\usepackage{booktabs}
\usepackage{soul}

\usepackage[pagebackref,breaklinks,colorlinks]{hyperref}

\usepackage[capitalize]{cleveref}
\crefname{section}{Sec.}{Secs.}
\Crefname{section}{Section}{Sections}
\Crefname{table}{Table}{Tables}
\crefname{table}{Tab.}{Tabs.}

\def\cvprPaperID{6} %
\def\confName{CVPR}
\def\confYear{2023}

\begin{document}

\title{Towards Self-Explainability of Deep Neural Networks with Heatmap Captioning and Large-Language Models}

\author{Osman Tursun, Simon Denman, Sridha Sridharan, and Clinton Fookes \\

\\
SAIVT Lab, Queensland University of Technology, Australia\\
{\{osman.tursun,s.denman,s.sridharan,c.fookes\}@qut.edu.au
}
}

\maketitle

\begin{abstract}
Heatmaps are widely used to interpret deep neural networks, particularly for computer vision tasks, and the heatmap-based explainable AI (XAI) techniques are a well-researched topic. However, most studies concentrate on enhancing the quality of the generated heatmap or discovering alternate heatmap generation techniques, and little effort has been devoted to making heatmap-based XAI automatic, interactive, scalable, and accessible. To address this gap, we propose a framework that includes two modules: (1) context modelling and (2) reasoning. We proposed a template-based image captioning approach for context modelling to create text-based contextual information from the heatmap and input data. The reasoning module leverages a large language model to provide explanations in combination with specialised knowledge. Our qualitative experiments demonstrate the effectiveness of our framework and heatmap captioning approach. The code for the proposed template-based heatmap captioning approach will be publicly available.

\end{abstract}

\section{Introduction}
\label{sec:intro}

Deep neural networks have continued to achieve very promising results for various machine learning and computer vision applications. However, they are commonly referred to as ``black-box" models, as their decision-making process lacks transparency and interpretability; though many attempts have been made to explain the hidden behaviour behind these models.

\begin{figure}[!h]
	
	\subfloat[Heatmaps and dog and cat captions]{\label{fig:dog_cat}\includegraphics[width=0.45\textwidth]{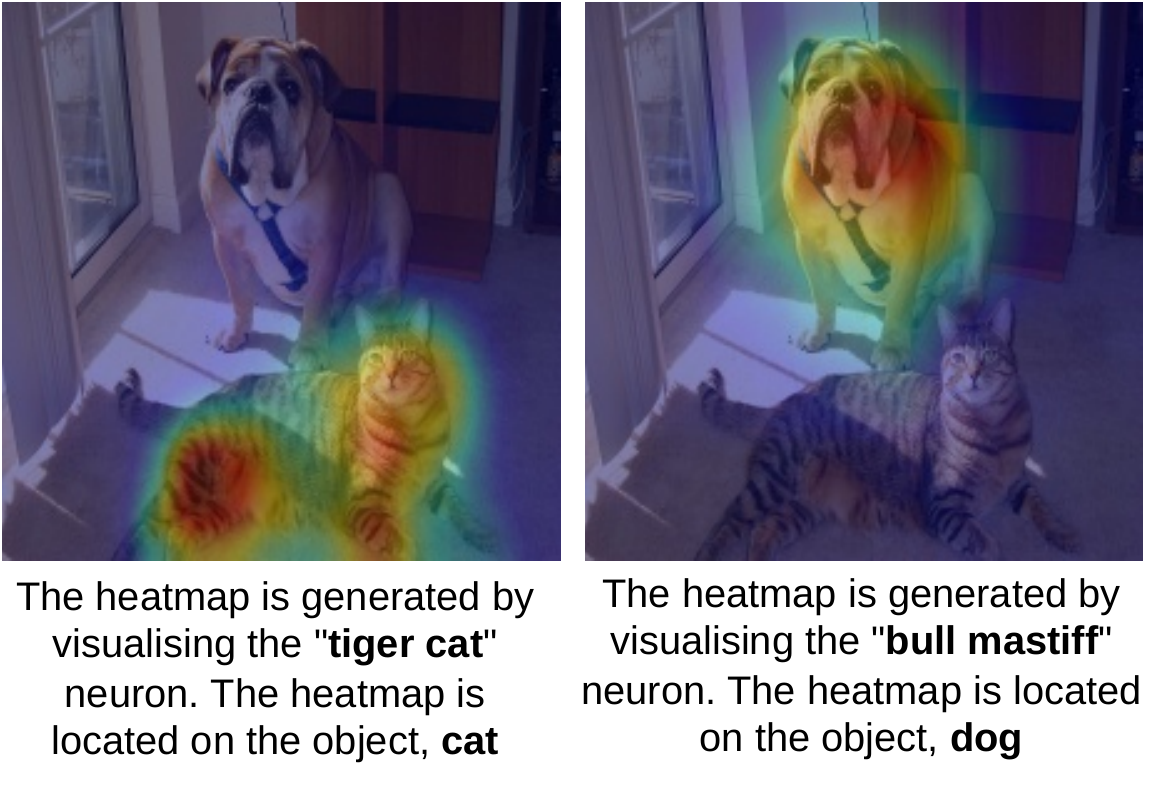}}

	\subfloat[An XAI report generated by ChatGPT. The captions of the generated heatmaps are sent to the ChatGPT as a prompt. ]{\label{fig:chatgpt}\includegraphics[width=0.45\textwidth]{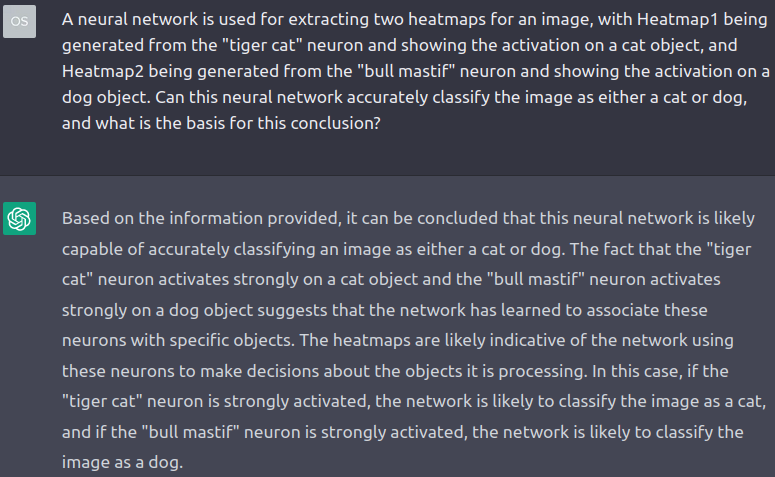}}
	\caption{A demonstration of the proposed framework for self-explainable heatmap-based XAI. In this example, ChatGPT is used for generating an XAI report.}
	\label{fig:concept}
\end{figure}

One widely studied interpretation approach is generating heatmaps to interrogate a neural network's decision. A heatmap is a graphical representation that highlights important elements of the data on which the neural network focuses for it's final decision. Heatmaps are useful for gaining insights into neural networks, and much research has been devoted to increasing the accuracy and improving the quality of heatmap visualisation \cite{tursun2022sess,zhang2021group,smilkov2017smoothgrad}. However, accurately interpreting heatmaps requires contextual knowledge and specialised knowledge of deep neural networks themselves, and without this information, any interpretation may be incomplete or misleading. Moreover, end-user accessibility, scalability and the automation of heatmap interpretation are limited. Therefore, here we focus on how to make heatmap-based explainable artificial intelligence (XAI) automatic, scalable, interpretable and interactive, so machine learning models will be more reliable and transparent across a range of applications and for users without specialist domain knowledge.

Our approach involves combining image captioning and with the power of large-language models such as GPT-3 \cite{brown2020language}, allowing us to extract critical task-related contextual information and easily access specialised expert knowledge. Both heatmap captioning and interaction with a large language model could be fully automatic, making the process scalable. The process can also be made interactive through user prompts based on the response of the large language model. For example, as shown in Figure \ref{fig:concept}, the heatmap captions are sent to ChatGPT \footnote{chat.openai.com}, which is developed based on GPT-3.5 \cite{ouyang2022training}. In this example, we asked ChatGPT to explain the capability of the neural network and comment on if it is able to separate the dog and cat classes. The chatbot generates a report which includes expert knowledge and is based on the specific context in this case, and the process is fully automatic. Based on needs, various general and technical questions can be asked. This makes heatmap-based XAI approachable, interactive, automatic and scalable. 

The primary objective of this study is on generating meaningful captions for heatmaps, which is not only the first step but also the most challenging part of our proposed framework. Although image captioning has achieved a marked improvement in recent years \cite{xu2015show, alayrac2022flamingo}, most methods are trained using natural image and text pairs in a supervised manner. To the best of our knowledge, there is no such dataset for heatmaps. Creating synthetic heatmap and text pairs is one potential approach, while generating synthetic captions for heatmaps is a non-trivial task as heatmaps have irregular shapes and positions in images. Moreover, heatmap captions take both source images and heatmap images into consideration. Therefore, existing approaches can't be directly applied to heatmap captioning. 

In this work, to address the aforementioned issues, we propose a simple template-based approach. The approach uses heatmaps to localise important regions in an image, and then extracts key attributes to generate captions. This approach is highly extendable as new attributes can be added to extract additional information from the heatmap and its source image.

Overall, the contribution of this work is summarised as: (i) We propose a fully-automatic framework for self-explainable heatmap-based XAI. This framework enables automatic, scalable and accessible XAI. (ii) We propose a template-based approach for captioning heatmaps, which is useful for extracting text-based contextual information from heatmaps. (iii) Through our qualitative experiments, we have demonstrated the potential of our framework and methodology.
\section{Related Studies}
\label{sec:lit}

In this study, our primary focus is on heatmap-based XAI approaches designed for black-box models such as neural networks. These approaches have gained popularity since the success of AlexNet, and many different techniques have been proposed for generating heatmaps to explain the behaviour of neural networks. These techniques are generally grouped into three categories: gradient-based \cite{simonyan2013deep,sundararajan2017axiomatic}, class activation-based\cite{zhou2016learning,selvaraju2017grad}, and perturbation-based \cite{fong2017interpretable,dabkowski2017real} methods.

Recently, the efficiency and quality of heatmap generation have significantly improved with the development of advanced techniques \cite{tursun2022sess, zhang2021group, smilkov2017smoothgrad,wang2020score,petsiuk2018rise}. Despite these advancements, expert interpretation is still required to fully understand the heatmaps and take the given task context into consideration. As Kim \etal \cite{kim2022help} discussed, a heatmap-based explanation is intuitive to high-AI background end-users, while not intuitive to low-AI background end-users. Heatmap captions and accessible specialised knowledge will reduce the gap between high-AI and low-AI end-users, making them more accessible as an interpretation method.

Human-in-the-loop XAI is an emerging area of research that focuses on incorporating human feedback into the XAI process. The proposed framework supports human feedback and interaction, making the XAI more approachable, consumable, and interactive. To address the importance of human feedback in XAI, Kim et al. \cite{kim2022hive} proposed a human-centred evaluation framework that takes into account the user's perspective in the evaluation process. Similarly, Lai et al. \cite{lai2023selective} presented a framework for generating selective explanations by leveraging human input, which enhances the understandability of AI explanations. These studies demonstrate the potential of human-in-the-loop XAI to improve the effectiveness and usability of XAI systems.

\section{Method}
\label{sec:meth}

\begin{figure}[!thb]
	\centering
	\includegraphics[width=0.5\textwidth]{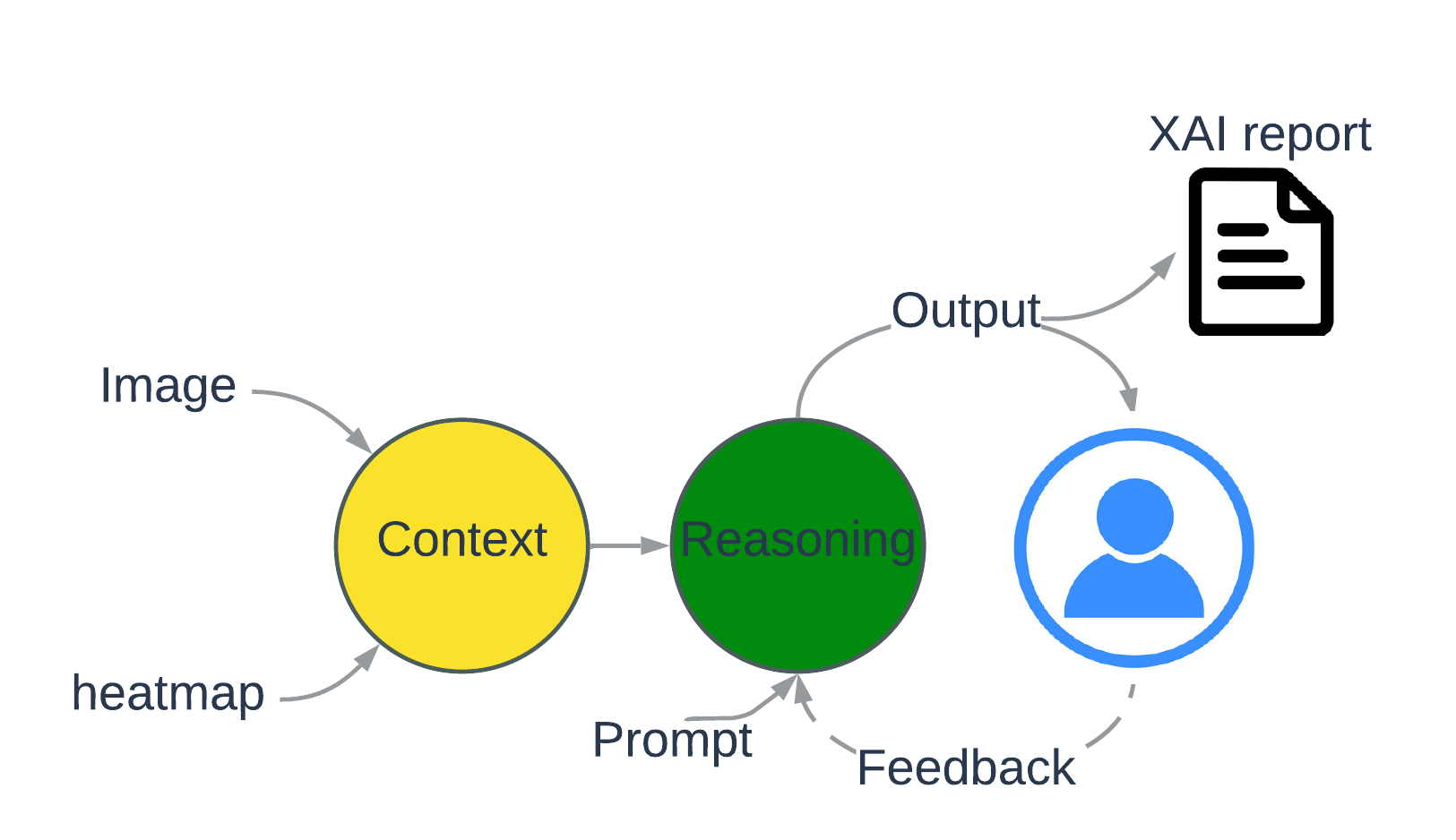}
	\caption{An illustration of the proposed framework. It consists of two main modules: \textit{context} and \textit{reasoning}. The context module generates contextual information through captioning heatmaps. The reasoning module leverages a large language model for analysing contextual information and user feedback in combination with specialised knowledge.}
	\label{fig:frame_diagram}
\end{figure}

In this work, we are proposing a new framework for scalable, self-interpretable, and interactive heatmap-based explainable artificial intelligence (XAI). As shown in Figure. \ref{fig:frame_diagram}, this framework consists of two main modules: \textit{context} and \textit{reasoning}. 

The context module extracts task-specific information from the input image and its corresponding heatmap(s), using the proposed heatmap captioning approach. The reasoning module leverages a large-language model to analyse this context information in combination with specialised expert knowledge. This process can be fully-automatic, making it highly scalable, but can also be made interactive by incorporating user feedback. For simplicity, we apply ChatGPT, the state-of-the-art large language model-based dialogue bot, for reasoning, while for the context module we will use the approach described in the following section.

\subsection{Captioning Heatmap with A Template-based Approach}

\begin{figure}
	\centering
	\includegraphics[width=0.5\textwidth]{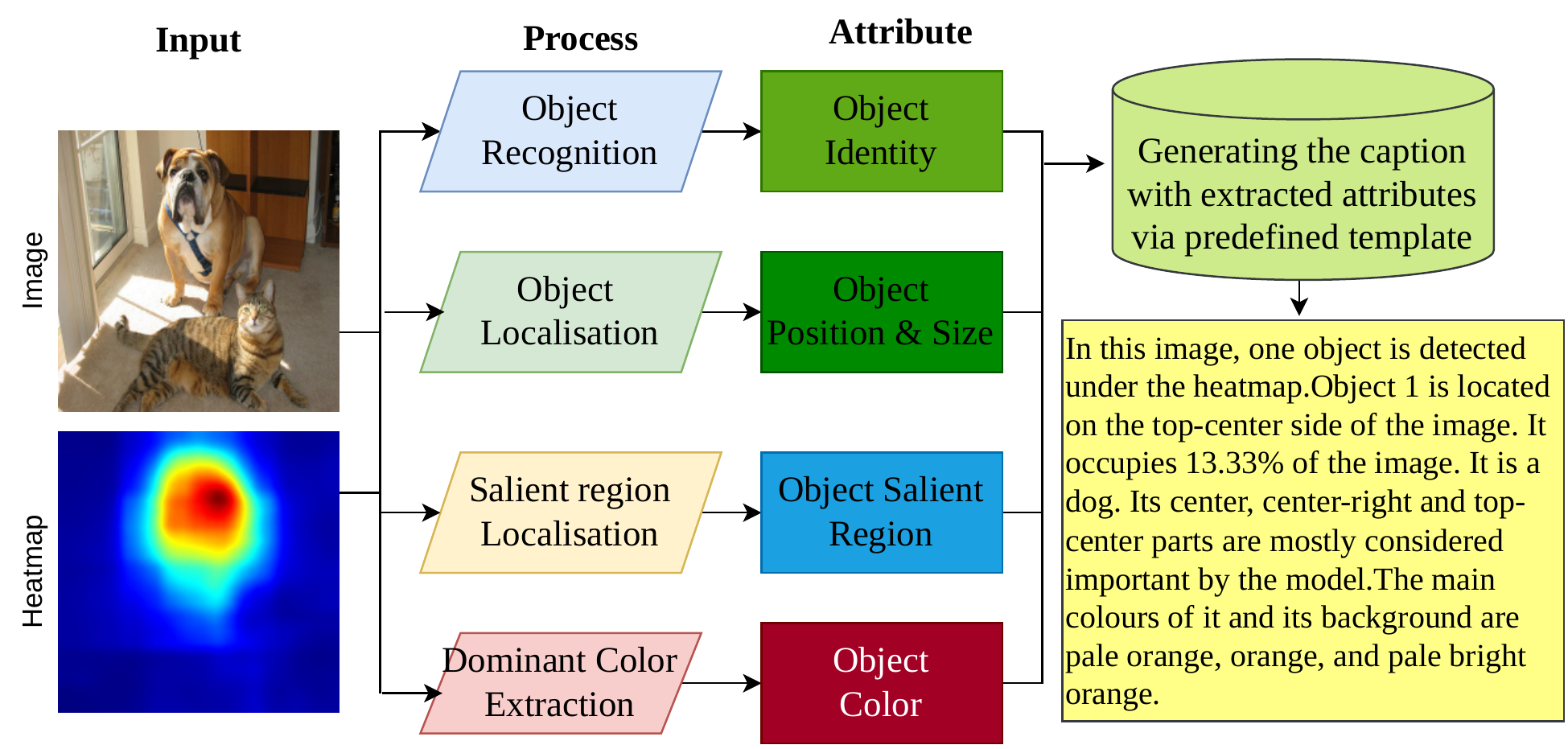}
	\caption{A diagram of the proposed template-based heatmap captioning method, which combines four different attributes to generate a heatmap caption.}
	\label{fig:diagram}
\end{figure}

To extract context information, captioning the heatmap image is very important. We, therefore, propose a template-based method to generate captions for heatmaps. Compared to mainstream deep learning-based end-to-end image captioning methods, template-based approaches do not require supervised training on image-text pairs. Another advantage of the template-based approach is its extendibility. Additional attributes can be added if required. As shown in Figure \ref{fig:diagram}, we extract four attributes from the objects located within the given heatmap. Here we briefly explain these attributes and the steps for extracting them.

To extract those four attributes, an image $I$ and its grayscale heatmap $H$ will be utilised. In this paper, SESS \cite{tursun2022sess} with GradCAM \cite{selvaraju2017grad} is chosen for extracting heatmaps. $H$ is used to localise objects and salient regions while $I$ is for the recognition and extraction of details. All objects under $H$ are localised by thresholding $H$ and applying connected component analysis. Each connected region is considered a single object and its rectangular bounding box is considered to be the bounding box of the object. Here, the notation $(x_i, y_i, w_i, h_i)$ represents the rectangle bounding box of the object $i$. For each located object, the following four attributes will be extracted.

The first attribute is the object identity. To identify the object, CLIP \cite{radford2021learning}, a language-image model, is applied as it has a strong zero-shot classification ability. For object $i$, the cropped region $I[x_i:x_i+w_i, y_i:y_i+h_i]$ is sent to CLIP for classification. In this study, a ViT-B/16 Transformer model and COCO classification labels \cite{lin2014microsoft} are used for classification.

The second attribute is the global position and size of the object. The centre of the bounding box of the object is used to localise the object. $I$ is equally divided into nine regions: top-left, centre-left, bottom-left, top-centre, centre, bottom-centre, top-right, centre-right and bottom-right. The region in which the object's centre $(x_i,y_i)$ is located is considered to be the object's position within the image. The relative size of the object relative to the image is also used included, which is equal to $(w_i*h_i)/size(I)$.

The third attribute is the salient regions of the object, where we seek to find the most important region of the object which contributes most to the decision of the model. For object $i$, $H[x_i:x_i+w_i, y_i:y_i+h_i]$ is equally divided into nine regions as we did for the global position attribute. The three regions with the highest mean intensity values are considered the most salient.  

The last attribute of the object specifies its dominant colour. This is determined using a colour naming algorithm applied to the HSV colour space, which is capable of naming the colour of a pixel with one of 93 semantic colour names, as described in \cite{liu2005region}. To identify the dominant colours for object $i$, the algorithm selects foreground pixels from the region $I[x_i:x_i+w_i, y_i:y_i+h_i]$ based on their heatmap intensity values. A pixel is from the foreground if its intensity exceeds 0.5. The algorithm then assigns a colour name to each selected pixel and calculates the percentage of each colour. The three colours with the highest percentage are then identified as the dominant colours for the object.

After the attribution extraction process is complete, the resulting attributes are inserted into predetermined templates to generate the final caption, as shown in Figure \ref{fig:diagram}.

\section{Experiment}
\label{sec:exp}

In this section, we provide some qualitative results including generated captions and XAI reports generated with ChatGPT. 

\begin{figure}[!tbh]
	\includegraphics[width=0.47\textwidth]{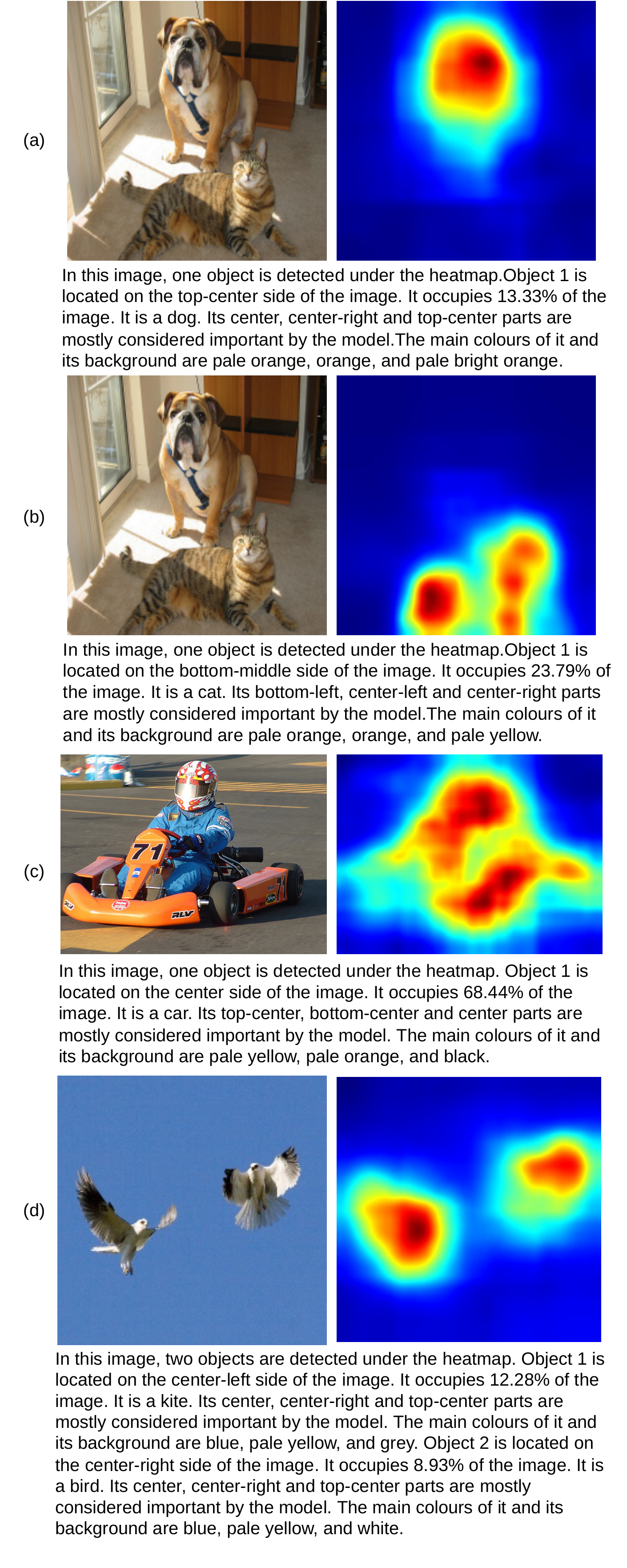}
	\caption{Qualitative results of generated captions. In each example, a caption is generated based on the image and its heatmap.}
	\label{fig:qual_caption}
\end{figure}

In Figure \ref{fig:qual_caption}, we display examples of the generated captions through the proposed template-based heatmap captioning approach. Those heatmaps are extracted from the ResNet50 \cite{he2016deep}. The proposed method can generate captions for both single-object scenes (a, b, c) and multiple-object scenes (d), covering all four key attributes. The generated captions are sensitive to changes in the heatmap, as demonstrated in (a-b). However, there are some specific instances where the generated captions are not entirely correct. For example, in (c), the heatmap covers both the "human driver" and "go-kart," but the caption only includes one object identified as a "car." Similarly, in (d), one object is labelled as "kite," while the other, which is identical, is labelled as "bird." These errors are due to the limitations of heatmap-based object localization and zero-shot classification, which can be minimised with the SOTA object localisation and classification techniques.

\subsection{Generating XAI Reports with ChatGPT}
To test out if the generated captions are informative enough for generating meaningful XAI report with a large-language model such as ChatGPT, we generated some reports with ChatGPT by sending generated captions to ChatGPT. We asked ChatGPT to answer questions including (i) Is a model working properly? (ii) What is the possible shortcoming of the model? (iii) Is a classification model able to locate certain types of objects?

When the correct captions and well-formed prompts are given, ChatGPT produces a well-written and informative XAI report. Examples of these reports are presented in the supplementary materials. The prompt for ChatGPT includes the following content:

\begin{itemize}
	\item How a heatmap is generated. For example, which neuron is used for extracting the heatmap.
	\item Heatmap caption(s). This is generated with the proposed template-based heatmap captioning approach, and are shown in Figure 4 of the main paper. However, we slightly modified the captions from Figures (c-d) to help obtain a reasonable answer from ChatGPT. 
	\item A question regarding the neural network. We underlined these questions in the following examples. We also underlined the key parts of ChatGPT's responses to those questions.
\end{itemize}

\noindent{\textbf{Prompt for Figure 4 (a-b)}}: \textit{A neural network is used for extracting two heatmaps for an image, with Heatmap1 being a generated form the "tiger cat" neuron and showing the activation on a cat object in the given image, and Heatmap 2 being generated from the "bull mastif" neuron and showing the activation on a dog object in the given object. Here are detailed information about heatmaps: Heatmap1: In this image, one object is detected under the heatmap. Object 1 is located on the top-center side of the image. It occupies 13.33\% of the image. It is a dog. Its center, center-right and top-center parts are mostly considered important by the model. The main colours of it and its background are pale orange, orange, and pale bright orange. Heatmap2:  In this image, one object is detected under the heatmap. Object 1 is located on the bottom-middle side of the image. It occupies 23.79\% of the image. It is a cat. Its bottom-left, center-left and center-right parts are mostly considered important by the model.The main colours of it and its background are pale orange, orange, and pale yellow.} \ul{\textit{Can this neural network accurately classify the image as either a cat or dog,  and what is the basis for this conclusion?}}

\noindent{\textbf{ChatGPT}}: \textit{Based on the information provided, the neural network has generated two heatmaps, one for detecting a cat and another for detecting a dog in the image. The description of Heatmap1 shows that the detected object is a dog, located on the top-center side of the image and occupying 13.33\% of the image. The network has identified the center, center-right, and top-center parts of the dog as the most important for its detection, and the main colors of the object and its background are pale orange, orange, and pale bright orange.
	On the other hand, the description of Heatmap2 shows that the detected object is a cat, located on the bottom-middle side of the image and occupying 23.79\% of the image. The network has identified the bottom-left, center-left, and center-right parts of the cat as the most important for its detection, and the main colors of the object and its background are pale orange, orange, and pale yellow.
	\ul{Since the network has generated separate heatmaps for detecting a cat and a dog in the image, and each heatmap shows a high degree of activation on the corresponding object, it is likely that the neural network is capable of accurately classifying the image as either a cat or a dog.}
	However, without additional information about the neural network's training data, architecture, and performance, it is difficult to make a definitive conclusion about its accuracy. It is possible that the network may have difficulty distinguishing between similar-looking objects or may not generalize well to other images. Further evaluation and testing would be necessary to determine the neural network's classification accuracy more precisely.}

\noindent{\textbf{Prompt for Figure 4 (c)}}: \textit{A neural network classified an image as "go-kart", and a heatmp is generated through visualising its most activated neuron. Here is the description of this heatmap: "In this image, one object is detected under the heatmap. Object 1 is located on the center side of the image. It occupies 68.44\% of the image. It is a go-kart with a human driver. Its top-center, bottom-center and center parts are mostly considered important by the model. The main colours of it and its background are pale yellow, pale orange, and black." \ul{What is the possible shortcoming of this neural network. hint: the human driver and the go-cart objects have the same degree of saliency.}}

\noindent{\textbf{ChatGPT}}: \textit{One possible shortcoming of this neural network is its lack of attention to distinguishing features that could differentiate the human driver from other objects or individuals. The fact that the model assigns the same degree of saliency to both the go-kart and the human driver suggests that the network may not be taking into account important details or features that could impact its classification.
	\\	
	For example, if the network is trained on a dataset of go-karts without human drivers, it may not have learned to differentiate between a go-kart with and without a driver. Alternatively, it is possible that the network's architecture or training data may not have been optimized to detect and classify human drivers accurately.
	\\	
	\ul{In either case, the lack of attention to distinguishing features could be a potential shortcoming of this neural network, as it may result in misclassifications or incorrect predictions when applied to other images.} To address this issue, the network could be trained on a more diverse dataset that includes images of go-karts with and without human drivers, as well as other types of vehicles and objects to ensure that it can accurately distinguish between them. Additionally, modifications to the network's architecture or training data may be necessary to improve its ability to identify and differentiate between relevant features.}
\\
\noindent{\textbf{Prompt for Figure 4 (d)}}: \textit{A neural network is used for extracting a heatmap for an image.  Detailed information of the heatmap: In this image, two objects are detected under the heatmap. Object 1 is located on the center-left side of the image. It occupies 12.28\% of the image. It is a bird. Its center, center-right and top-center parts are mostly considered important by the model. The main colours of it and its background are blue, pale yellow, and grey. Object 2 is located on the center-right side of the image. It occupies 8.93\% of the image. It is a bird. Its center, center-right and top-center parts are mostly considered important by the model. The main colours of it and its background are blue, pale yellow, and white. \ul{Based on the heatmap information, is this network useful for locating a bird object?}}

\noindent{\textbf{ChatGPT}}: \textit{Based on the information provided, it is likely that the neural network is capable of accurately detecting and locating bird objects in an image. The heatmap generated by the network indicates that two bird objects were detected in the image, each occupying a distinct region of the image and identified by specific regions of importance, including the center, center-right, and top-center parts of each object.
Furthermore, the description of the main colors of each object and its background suggests that the network is capable of identifying and distinguishing between different color patterns associated with bird objects.
\ul{Overall, the heatmap information provided suggests that the neural network is useful for locating bird objects in images, particularly those with distinct color and feature patterns that are easily distinguishable by the network.} However, it is important to note that the accuracy and effectiveness of the network may vary depending on the quality and diversity of the training data and the complexity and adaptability of the network's architecture.}

\section{Future Work and Conclusion}
\label{sec:con}

In this study, we present a promising framework for automatic, interactive, and scalable XAI report generation using a large-language model. Specifically, we utilized a proposed template-based heatmap caption generation approach to provide contextual information for a large-language-based reasoning module, such as ChatGPT. Our results demonstrate the promise of this approach, yet also highlight the importance of an accurate template-based captioning approach, and we note that further improvements are required to realise a fully automatic XAI report generation. We observe that while the generated captions were informative, they lack diversity and can contain redundant information. Future research could explore the use of deep learning-based image captioning approaches to address these limitations. Furthermore, we found that the reports generated by ChatGPT were not concise and required well-designed prompts. Despite these shortcomings, designing a large-language model specifically for XAI report generation shows promise as a research direction.
\\

{\small
\bibliographystyle{ieee_fullname}
\bibliography{egbib}

\begin{thebibliography}{10}\itemsep=-1pt

\bibitem{alayrac2022flamingo}
Jean-Baptiste Alayrac, Jeff Donahue, Pauline Luc, Antoine Miech, Iain Barr,
  Yana Hasson, Karel Lenc, Arthur Mensch, Katie Millican, Malcolm Reynolds,
  et~al.
\newblock Flamingo: a visual language model for few-shot learning.
\newblock {\em arXiv preprint arXiv:2204.14198}, 2022.

\bibitem{brown2020language}
Tom Brown, Benjamin Mann, Nick Ryder, Melanie Subbiah, Jared~D Kaplan, Prafulla
  Dhariwal, Arvind Neelakantan, Pranav Shyam, Girish Sastry, Amanda Askell,
  et~al.
\newblock Language models are few-shot learners.
\newblock {\em Advances in neural information processing systems},
  33:1877--1901, 2020.

\bibitem{dabkowski2017real}
Piotr Dabkowski and Yarin Gal.
\newblock Real time image saliency for black box classifiers.
\newblock {\em Advances in neural information processing systems}, 30, 2017.

\bibitem{fong2017interpretable}
Ruth~C Fong and Andrea Vedaldi.
\newblock Interpretable explanations of black boxes by meaningful perturbation.
\newblock In {\em Proceedings of the IEEE international conference on computer
  vision}, pages 3429--3437, 2017.

\bibitem{he2016deep}
Kaiming He, Xiangyu Zhang, Shaoqing Ren, and Jian Sun.
\newblock Deep residual learning for image recognition.
\newblock In {\em Proceedings of the IEEE conference on computer vision and
  pattern recognition}, pages 770--778, 2016.

\bibitem{kim2022hive}
Sunnie~SY Kim, Nicole Meister, Vikram~V Ramaswamy, Ruth Fong, and Olga
  Russakovsky.
\newblock Hive: evaluating the human interpretability of visual explanations.
\newblock In {\em Computer Vision--ECCV 2022: 17th European Conference, Tel
  Aviv, Israel, October 23--27, 2022, Proceedings, Part XII}, pages 280--298.
  Springer, 2022.

\bibitem{kim2022help}
Sunnie~SY Kim, Elizabeth~Anne Watkins, Olga Russakovsky, Ruth Fong, and
  Andr{\'e}s Monroy-Hern{\'a}ndez.
\newblock " help me help the ai": Understanding how explainability can support
  human-ai interaction.
\newblock {\em arXiv preprint arXiv:2210.03735}, 2022.

\bibitem{lai2023selective}
Vivian Lai, Yiming Zhang, Chacha Chen, Q~Vera Liao, and Chenhao Tan.
\newblock Selective explanations: Leveraging human input to align explainable
  ai.
\newblock {\em arXiv preprint arXiv:2301.09656}, 2023.

\bibitem{lin2014microsoft}
Tsung-Yi Lin, Michael Maire, Serge Belongie, James Hays, Pietro Perona, Deva
  Ramanan, Piotr Doll{\'a}r, and C~Lawrence Zitnick.
\newblock Microsoft coco: Common objects in context.
\newblock In {\em European conference on computer vision}, pages 740--755.
  Springer, 2014.

\bibitem{liu2005region}
Ying Liu, Dengsheng Zhang, Guojun Lu, and Wei-Ying Ma.
\newblock Region-based image retrieval with high-level semantic color names.
\newblock In {\em 11th International Multimedia Modelling Conference}, pages
  180--187. IEEE, 2005.

\bibitem{ouyang2022training}
Long Ouyang, Jeff Wu, Xu Jiang, Diogo Almeida, Carroll~L Wainwright, Pamela
  Mishkin, Chong Zhang, Sandhini Agarwal, Katarina Slama, Alex Ray, et~al.
\newblock Training language models to follow instructions with human feedback.
\newblock {\em arXiv preprint arXiv:2203.02155}, 2022.

\bibitem{petsiuk2018rise}
Vitali Petsiuk, Abir Das, and Kate Saenko.
\newblock Rise: Randomized input sampling for explanation of black-box models.
\newblock {\em arXiv preprint arXiv:1806.07421}, 2018.

\bibitem{radford2021learning}
Alec Radford, Jong~Wook Kim, Chris Hallacy, Aditya Ramesh, Gabriel Goh,
  Sandhini Agarwal, Girish Sastry, Amanda Askell, Pamela Mishkin, Jack Clark,
  et~al.
\newblock Learning transferable visual models from natural language
  supervision.
\newblock In {\em International conference on machine learning}, pages
  8748--8763. PMLR, 2021.

\bibitem{selvaraju2017grad}
Ramprasaath~R Selvaraju, Michael Cogswell, Abhishek Das, Ramakrishna Vedantam,
  Devi Parikh, and Dhruv Batra.
\newblock Grad-cam: Visual explanations from deep networks via gradient-based
  localization.
\newblock In {\em Proceedings of the IEEE international conference on computer
  vision}, pages 618--626, 2017.

\bibitem{simonyan2013deep}
Karen Simonyan, Andrea Vedaldi, and Andrew Zisserman.
\newblock Deep inside convolutional networks: Visualising image classification
  models and saliency maps.
\newblock {\em arXiv preprint arXiv:1312.6034}, 2013.

\bibitem{smilkov2017smoothgrad}
Daniel Smilkov, Nikhil Thorat, Been Kim, Fernanda Vi{\'e}gas, and Martin
  Wattenberg.
\newblock Smoothgrad: removing noise by adding noise.
\newblock {\em arXiv preprint arXiv:1706.03825}, 2017.

\bibitem{sundararajan2017axiomatic}
Mukund Sundararajan, Ankur Taly, and Qiqi Yan.
\newblock Axiomatic attribution for deep networks.
\newblock In {\em International conference on machine learning}, pages
  3319--3328. PMLR, 2017.

\bibitem{tursun2022sess}
Osman Tursun, Simon Denman, Sridha Sridharan, and Clinton Fookes.
\newblock Sess: Saliency enhancing with scaling and sliding.
\newblock In {\em Computer Vision--ECCV 2022: 17th European Conference, Tel
  Aviv, Israel, October 23--27, 2022, Proceedings, Part XII}, pages 318--333.
  Springer, 2022.

\bibitem{wang2020score}
Haofan Wang, Zifan Wang, Mengnan Du, Fan Yang, Zijian Zhang, Sirui Ding, Piotr
  Mardziel, and Xia Hu.
\newblock Score-cam: Score-weighted visual explanations for convolutional
  neural networks.
\newblock In {\em Proceedings of the IEEE/CVF conference on computer vision and
  pattern recognition workshops}, pages 24--25, 2020.

\bibitem{xu2015show}
Kelvin Xu, Jimmy Ba, Ryan Kiros, Kyunghyun Cho, Aaron Courville, Ruslan
  Salakhudinov, Rich Zemel, and Yoshua Bengio.
\newblock Show, attend and tell: Neural image caption generation with visual
  attention.
\newblock In {\em International conference on machine learning}, pages
  2048--2057. PMLR, 2015.

\bibitem{zhang2021group}
Qinglong Zhang, Lu Rao, and Yubin Yang.
\newblock Group-cam: Group score-weighted visual explanations for deep
  convolutional networks.
\newblock {\em arXiv preprint arXiv:2103.13859}, 2021.

\bibitem{zhou2016learning}
Bolei Zhou, Aditya Khosla, Agata Lapedriza, Aude Oliva, and Antonio Torralba.
\newblock Learning deep features for discriminative localization.
\newblock In {\em Proceedings of the IEEE conference on computer vision and
  pattern recognition}, pages 2921--2929, 2016.

\end{thebibliography}


\begin{thebibliography}{}\itemsep=-1pt

\end{thebibliography}
}

\end{document}